\documentclass[conference]{IEEEtran}
\IEEEoverridecommandlockouts
\usepackage{cite}
\usepackage{amsmath,amssymb,amsfonts}
\usepackage{algorithmic}
\usepackage{graphicx}
\usepackage{textcomp}
\usepackage{xcolor}

\usepackage{times}
\usepackage{latexsym}
\usepackage{fixltx2e}
\usepackage{balance}

\usepackage{comment}

\usepackage{hyperref}
\usepackage{subcaption}
\usepackage{url}
\usepackage{graphicx}
\usepackage{amsmath}
\DeclareMathOperator*{\argmax}{\arg\!\max}

\def\BibTeX{{\rm B\kern-.05em{\sc i\kern-.025em b}\kern-.08em
    T\kern-.1667em\lower.7ex\hbox{E}\kern-.125emX}}
    
\begin{document}

\title{Exploring Sentence Vector Spaces through Automatic Summarization\\
%{\footnotesize \textsuperscript{*}Note: Sub-titles are not captured in Xplore and
%should not be used}
\thanks{National Science Foundation, Award No. 1359275}
}

\author{\IEEEauthorblockN{Adly Templeton
}
\IEEEauthorblockA{\textit{Department of Computer Science
}\\
\textit{Williams College
}\\
880 Main St, Williamstown, MA 01267
 \\
at7@williams.edu
}
\and
\IEEEauthorblockN{Jugal Kalita
}
\textit{Department of Computer Science
} \\
\textit{University of Colorado, Colorado Springs 
}\\
jkalita@uccs.edu
}

\maketitle

\renewcommand\floatpagefraction{.9}
\renewcommand\dblfloatpagefraction{.9} % for two column documents
\renewcommand\topfraction{.9}
\renewcommand\dbltopfraction{.9} % for two column documents
\renewcommand\bottomfraction{.9}
\renewcommand\textfraction{.1}   
\setcounter{totalnumber}{50}
\setcounter{topnumber}{50}
\setcounter{bottomnumber}{50}

%\usepackage{url}

%\setlength\titlebox{5cm}
% You can expand the titlebox if you need extra space
% to show all the authors. Please do not make the titlebox
% smaller than 5cm (the original size); we will check this
% in the camera-ready version and ask you to change it back.

%\newcommand\BibTeX{B{\sc ib}\TeX}

\date{}

\newcommand{\fix}{\marginpar{FIX}}
\newcommand{\new}{\marginpar{NEW}}

%\begin{document}

\begin{abstract}
Given vector representations for individual words, it is  necessary to compute vector representations of sentences for many applications in a compositional manner, often using artificial neural networks.
 Relatively little work has explored the internal structure and properties  of such sentence vectors. In this paper, we explore the properties of sentence vectors in the context of automatic summarization. In particular, we show that cosine similarity between sentence vectors and document vectors is strongly correlated with sentence importance and that vector semantics can identify and correct gaps between the sentences chosen so far and the document. In addition, we identify specific dimensions which are linked to effective summaries. To our knowledge, this is the first time specific dimensions of sentence embeddings have been connected to sentence properties. We also compare the features of different methods of sentence embeddings. Many of these insights have applications in uses of sentence embeddings far beyond summarization.
\end{abstract}

%\[ {\overrightarrow{x}_{i+1}} = \overrightarrow{x}_i - H^{-1} (\overrightarrow{x}_i) \;  \bigtriangledown f (\overrightarrow{x}_i) \]

\section{Introduction}
 Vector semantics  represent words as vectors in a high-dimensional space, where vectors which are close to each other  supposedly have similar meanings. 
 Various models of vector semantics, such as LSA \cite{la97}, word2vec \cite{mi13}, and GLOVE \cite{pe14} have proved to be successful in many natural language processing applications.

Given word vectors, it is often necessary to compute vectors for sentences or documents. 
A number of techniques for sentence embeddings have been proposed. For example, \textit{paragraph vectors} (also known as doc2vec) \cite{le14} resemble word2vec vectors, except that a classifier uses an additional `paragraph vector' to predict words in a \textit{skip-gram} model. 
%When trained, the paragraph vectors capture the meanings of their paragraphs.
\textit{skip-thought vectors} use an encoder-decoder neural network which, given a sentence, predicts the surrounding sentences \cite{ki15}. 
In spite of the popularity of neural networks to obtain sentence vectors, approaches based on linear combinations of the word vectors have managed to achieve state-of-the-art results for non-domain-specific tasks \cite{wi15}. In particular, Arora et al. \cite{ar16}  find  the weighted average of the word vectors (less frequent words weighted higher). Note that this weighting is roughly analogous to weighting by TF-IDF. The second step is to subtract the projection along a specific vector, called the ``common discourse vector", correlated with words such as ``the" or ``and" which appear consistently in all English contexts. This vector is found by taking the first principal component of a representative sample of text vectors. 
Conneau et al. \cite{Conneau2017} show that a bi-directional LSTM network with max pooling produces the most effective sentence representation, considering a number of tasks. 

This paper explores the space of sentence vectors  in order to understand how sentences picked by several extractive summarization algorithms relate to document vectors.
% It also studies how the dimensionality of sentence embeddings impacts on sentence properties. 
Section 2 introduces the main topics of the paper, followed by related work in Section 3. Section 4 discusses the combinations of vector functions and sentence selection functions we use to explore how they affect one another, followed by a detailed analysis in Section 5. Section 6 concludes the paper. 

\section{Extractive Summarization and Sentence Vectors}
%The large volume of news articles published every day motivates effective forms of automatic summarization. 
Extractive summarization performs sentence extraction, in which entire unmodified sentences are selected from the original document, and then concatenated to form a short summary, which ideally contains the most important information from the original document.
Sentence extraction models use various metrics to select sentences. 
 %based on the occurrence of individual words. 
For instance, a state-of-the-art graph-based summarization model  uses the words shared between sentences for all three of their major metrics: importance, redundancy, and coherence \cite{pa15}. 
%However metrics based on the occurrence of specific words, and not the meaning of these words, may perform sub-optimally in certain situations (such as dealing with synonyms).

In this paper, we consider extractive summarization as a task that is essentially reducible to \textit{sentence selection}. We want to select a subset of sentences from the set of all sentences in the original document, which maximizes the quality of the summary while remaining under some word limit. Note that the quality of each individual sentence depends of the other sentences in the summary. In particular, a good summary should contain minimal redundancy. This trade-off between sentence salience and redundancy is the basis for many summarization algorithms.

\begin{comment}
Although we are primarily concerned with multi-document summarization algorithms, we 
%which summarize a \textit{cluster} of related documents. However, all our algorithms 
disregard the document-by-document information and work with the the documents as clusters.
In addition, 
a practical summarization system  likely includes other steps after sentence selection, such as \textit{sentence reordering} or \textit{text-simplification}. We do not consider these issues in this paper. 
\end{comment}

Sentence vectors, in some form,  are necessary for dealing with sentence-level or higher-level constructs such as paragraphs and documents. Sentence vectors may be helpful in comparing sentences for tasks such as computing sentence similarity, question answering, sentiment analysis  and document classification. Each of these tasks is likely to focus, implicitly or explicitly, on some unique aspects of sentence vectors. For example, sentence similarity computation may need to consider the sequencing of the words for a finer modeling of the sentence compared to what is necessary for multi-document summarization. 
In this paper, we look at summarization only. 
We believe that the nature of summarization requires some specific properties from sentence vectors, making it a suitable example for which current sentence embedding methods can be easily used. We assume sentence vectors are computed explicitly for use in summarization and 
 compare the effectiveness of different sentence embeddings in practice. 

\section{Related Work}
State-of-the-art extractive summarization techniques have been achieved with a wide variety of methods. 
K{\aa}geb{\"a}ck et al.  \cite{ka14} used cosine similarity between the sum of word2vec embeddings, as well as a recursive auto-encoder, when modifying  an existing summarization algorithm. Cao et al. \cite{ca15} used a recursive neural network (RNN), which operates on a parse tree of a sentence, to rank sentences for summarization.
Cheng and Lapata \cite{ch16}  used an RNN sentence extractor with an attention mechanism to perform document encodings and considered the previously selected sentences and the current sentence in making decisions.
Ren et al. \cite{re16} achieved state-of-the-art results through a regression-based approach, using a variety of engineered features, including the  Average Word Embedding, to estimate the redundancy-aware importance of a sentence given the sentences already selected. 
 Cheng and Lapata \cite{ch16} used word embeddings as the input to a neural network as part of summarization, but they did not directly compare embeddings. They used a single-layer convolutional neural network to encode sentences, and a LSTM neural network to encode documents.
Nayeem and Chali \cite{na17} recently achieved state-of-the-art results using a modified version of LexRank, using a combination of cosine similarity of weighted averages of vectors and named entity overlap.
 
To our knowledge, no one has explored the direct use of sentence embedding methods such as Paragraph Vectors or Skip-Thought vectors in summarization, although some work has been done on summarization using  representations  for individual words.

\section{Methods}
There are many potential ways to use vector semantics. To explore the design space, we consider combinations of \textit{vector functions} and \textit{selector functions}, functions which, given vector representations for sentences, extracts a summary. 

\subsection{Vector Functions}
We use the following vector functions. All sentence embeddings are normalized to produce unit vectors.

%\vspace*{-2pt}
%\begin{itemize}
%\setlength{\itemsep}{-2pt}
%\item 
\textit{SIF Average}: The most basic sentence embedding is simply the weighted average of word vectors from \cite{ar16}, without the common component removal. We use the Brown corpus \cite{fr79} for word frequency information. 

%\item 
\textit{Arora}: This method  \cite{ar16} is equivalent to the one above, except with common component removal added. We use the Brown corpus both to compute the common component vector, and for word frequency information. 

%\item 
\textit{Paragraph Vectors}: It is the paragraph vector approach described above. We used the 300-dimensional DBOW model pretrained by \cite{la16} on the Wikipedia corpus.

%\item 
\textit{Skip-Thought Vectors} It is the skip-thought vector approach described above. We used the 4800-dimensional combined-skip model \cite{ki15,al16}.

%\end{itemize}

\subsection{Selector Functions}
We use a large variety of  selector functions  to explore interaction effects between selector functions and vector functions.

%\vspace*{-2pt}
%\begin{itemize}
%\setlength{\itemsep}{-2pt}
%\item 

{\em Random}: This selector simply selects sentences at random, until the word limit is reached. This provides a lower-bound on the performance of an effective algorithm.

%\item 
\textit{Near}: \textit{Near} selects the sentences whose sentence vectors have the highest cosine similarity with the document vector.

%\item 
\textit{Near Non-redundant}:  To balance redundancy with salience, {\em Near Non-redundant} down-weights the cosine similarity scores by their average cosine similarity the sentences selected so far. Because this redundancy measure is strongly (quadratically) correlated with cosine similarity to the document, we fit a regression for redundancy for each vector function, and use the residual on this regression for the final algorithm.

%\item 
\textit{Greedy}: The greedy selector, at each step, selects the sentence that maximizes the cosine similarity of the new summary (including previously selected sentences). This is subtly different from the Near selector for average-based vector functions, but significantly different for Paragraph Vectors.

%\item
 \textit{Brute Force}: Another attempt at optimizing the cosine similarity between the summary and the document, this selector creates a pool of the 20 sentences with the highest cosine similarity. From this pool, every combination of sentences (with an appropriate word count) is tried, and the combination with the highest cosine similarity is selected as the summary.

%\item 
\textit{Max Similarity}: A  selector which computes results for both Greedy and Brute Force selectors and then selects the result with the highest cosine similarity to the document vector.

%\item 
\textit{Near-then-Redundancy}: Similar to the Brute Force selector, this selector creates the same pool of sentences, except with a size of 15. From this pool, this algorithm optimizes via brute force to minimize redundancy, defined as the average cosine similarity between pairs of sentences. The size of the sentence pool, which is essentially a  shortcut in the Brute Force selector, is a performance-critical hyper-parameter.

%\item 
\textit{Cluster}: We use an agglomerative clustering algorithm, using cosine similarity as its distance metric, to find clusters in the set of sentence embeddings. We then find the sentence closest to the average of each cluster and add it to the summary. To ensure we find summaries which meet the word-length requirement, we increase the number of clusters we search for until we have selected sentences totaling 100 words.

%\item 
\textit{PCA}: This selector performs Principal Component Analysis (PCA) \cite{Wold1987} on the set of sentence vectors in a document. Then, the algorithm selects  one sentence closest to the first component, one sentence closest to the second component, and so on, until the length capacity is met. 

%\item 
\textit{LexRank}: This selector is based on the classical LexRank algorithm \cite{er04}. It builds a weighted graph where nodes represent sentences, and weights are the similarities between sentences, as determined by cosine similarity between TF-IDF vectors. The PageRank algorithm \cite{Page1999} is then used on this graph to identify the most salient sentences for extraction. In our algorithm,  the weights of edges are determined by the cosine similarity between sentence embeddings.

%\end{itemize}

\section{Performance of Selector Functions and Detailed Analysis}
\begin{table*}[h]
\centering
\begin{tabular}{|l|llll|}
\hline
                                       & \textbf{SIF Average} & \textbf{Arora}      & \textbf{Paragraph Vectors} & \textbf{Skipthought}     \\ \hline
                                       
\textbf{LexRank}                       & 32.6 (6.8)           & 32.6 (6.8)          & 32.6 (6.8)                 & {32.6 (6.8)}      \\ 
\textbf{Near Nonredundant}             & 33.6 (6.1)           & {34.5} (6.3) & 32.6 (5.5)                 & 32.1 (4.9)               \\

\textbf{Brute Force}                   & 32.0 (5.7)           & 32.2 (6.3)          & {33.0 (6.6)}        & 31.4 (4.5)              \\ 
\textbf{Near-then-Redundancy}          & 33.2 (6.2)           & 34.2 {(6.9)} & 31.5 (5.4)                 & 33.1(5.3)             \\ 
\textbf{PCA}                           & 32.9 (5.6)           & 33.5 (5.6)          & 32.0 (5.5)                 & NA               \\ 
\textbf{Max Similarity}                & 32.0 (5.7)           & 32.2 (6.3)          & {33.0 (6.6)}        & NA  \\ 

\textbf{Greedy}                        & {35.1 (7.0)}  & 33.1 (6.0)          & NA                  & NA                \\ 

\textbf{Near}                          & 32.5 (5.4)           & 32.2 (5.5)          & 33.1 (6.1)                 & NA               \\
\textbf{Cluster}                       & NA       & NA           & NA                  & 32.1 (4.6)              \\ \hline
\end{tabular}
\caption{ROUGE-1 Results on the DUC 2004 dataset. ROUGE-2 results in parentheses. All combinations which do not perform significantly better than random chance ($p < .05$, using a paired t-test) are replaced with `NA' for clarity. SIF Average with either Max Similarity or Brute Force were included, despite having p=.051. In addition, one combination (Max Similarity with Skipthought Vectors) are not computed, but are not expected to perform better than chance. Selector Functions are roughly organized according to the vector functions with which they are effective. For Skipthought vectors, \textit{docvec-avg} is used.}
\label{tab:results}
\end{table*}

\begin{table*}[h]
\centering
\begin{tabular}{|l|llll|}
\hline
\textbf{}                     & \textbf{SIF Average} & \textbf{Arora} & \textbf{Paragraph Vectors} & \textbf{Skipthought}                            \\
\hline
\textbf{Near Nonredundant}    & -1.98                & -1.72          & -0.734                     & +3.81                                           \\
\textbf{Brute Force}          & -0.323               & -0.256         & -1.24                      & +3.42                                           \\
\textbf{Near-then-Redundancy} & +0.739               & -0.584         & -0.205                     & +3.46                                           \\
\textbf{Max Similarity}       & -0.323               & -0.256         & -1.24                      & NA \\

\textbf{Greedy}               & +0.254               & -0.813         & -3.11                      & NA \\

\textbf{Near}                 & -0.868               & -0.0652        & -5.53                      & +1.76                                           \\
\textbf{Total Average}        & -.417                & -.614          & -2.01                      & +2.74\\
\hline
\end{tabular}

\caption{A comparison of document vector methods. Numbers represent the difference in ROUGE-1 scores between document vector methods. Positive numbers represent a gain when using \textit{docvec-avg}. Selectors which do not use the document vector have been omitted.}
\label{tab:docvec}
\end{table*}

%\subsection{Experimental Evaluation}
The purpose of our experiments is to combine every vector function with every selector function, use the combination to perform extractive text summarization, and then conduct an in-depth analysis to shed light on the performance and effectiveness of various types sentence vectors. 

Despite the small sample size, we use the standard DUC datasets to compare our results with other summarization algorithms. We split the document clusters in the DUC 2004 dataset into a testing set and a validation set of approximately equal sizes. The pre-defined training set of the DUC 2001 dataset was used as a training set  for some of the graphs and data analysis presented here.
ROUGE  \cite{li04} has been shown to correlate strongly with human summarization judgments \cite{ra13}, and is our primary metric for evaluating summaries\footnote{We trucate summaries to 100 words and use the following parameters, for direct comparison with Hong et al. \cite{ho14}:  -n 4 -m -a -l 100 -x -c 95 -r 1000 -f A -p 0.5 -t 0.}. We report ROUGE-1 and ROUGE-2 statistics, which correspond to unigrams and bigrams, respectively.
We present results for Multi-Document Summarization on the DUC 2004 dataset (Table \ref{tab:results}). ROUGE scores are presented to facilitate the analysis given below, and not to anoint one particular combination of vector and selector functions as the best summarizer.

\subsection{Analysis and Discussion}
An analysis of the underlying data provides  useful insights into the behavior of vector semantics in computational tasks.

\paragraph{Distributions of Cosine Scores}

\begin{figure}
\begin{subfigure}{0.49\textwidth}
    \centering
       \includegraphics[width=100pt]{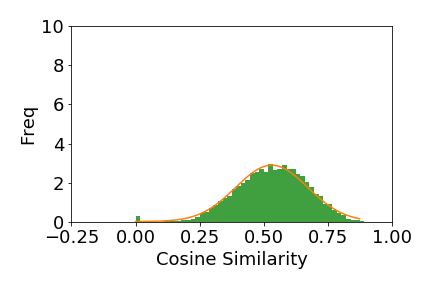}
        \includegraphics[width=100pt]{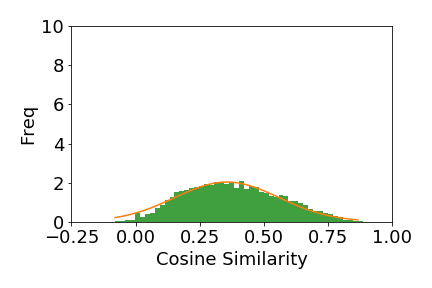}
       \caption{SIF Average amd Arora}
\end{subfigure}
%\begin{subfigure}{0.4\textwidth}
%    \centering
%        \includegraphics[width=100pt]{fig/dist_Arora.png}
%        \caption{Arora}
%\end{subfigure}
\begin{subfigure}{0.49\textwidth}
    \centering
        \includegraphics[width=100pt]{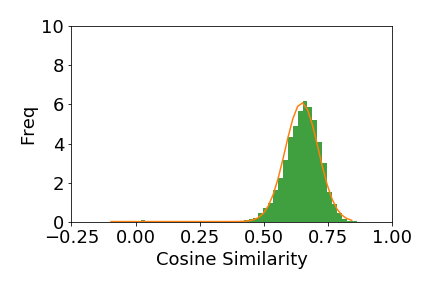}
        \includegraphics[width=100pt]{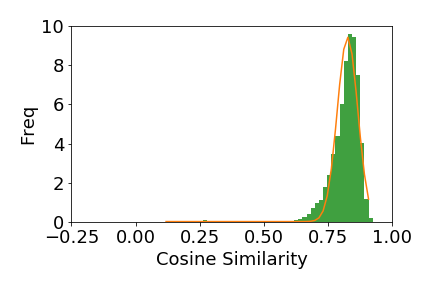}
        \caption{Paragraph Vectors and Skipthought}
        \end{subfigure}
%\begin{subfigure}{0.4\textwidth}
%    \centering
%        \includegraphics[width=100pt]{fig/dist_Skipthought.png}
 %       \caption{Skipthought}
%\end{subfigure}
\caption{Distribution of cosine similarity scores between each sentence vector and their corresponding document vector, for all four vector functions.}
\label{fig:dist}
\end{figure}

The cosine scores between all sentence vectors and the corresponding document vectors follow a normal distribution for all vector functions (Fig. \ref{fig:dist}), but this effect is most pronounced for paragraph vectors ($r^2=.996$). In addition, the sentence embeddings for paragraph vectors and skip-thought vectors are far closer to the document embedding than would be expected from a random distribution, with mean cosine similarities of .65 and .84, respectively. Unsurprisingly, this also holds for Average and Arora, though the similarity is notably lower (.52 and .35, respectively).

\paragraph{Correlation of Cosine Scores with Good Summaries}
\begin{figure}[h]
\includegraphics[width=4cm]{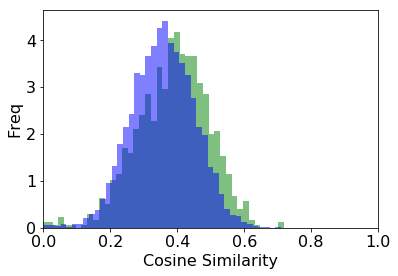}
\includegraphics[width=4cm]{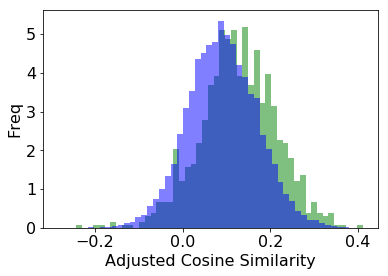}
\centering
\caption{Cosine Similarity to the Document Vector for non-optimal (blue) and optimal (green) sentences. Figure on the right shows Cosine Similarity adjusted for sentence word count.}
\label{fig:dualcos}
\end{figure}

\begin{figure}
\begin{subfigure}{0.49\textwidth}
    \centering
       \includegraphics[width=100pt]{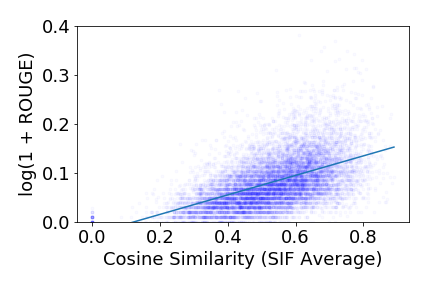}
        \includegraphics[width=100pt]{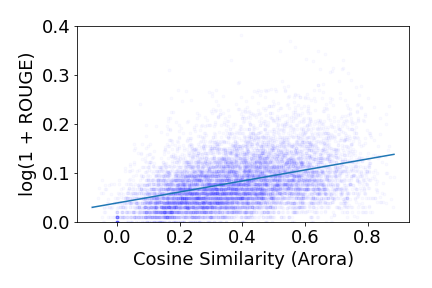}
       \caption{SIF Average (r=.53) and Arora (r=.40)}
\end{subfigure}
%\begin{subfigure}{0.49\textwidth}
%    \centering
%        \includegraphics[width=150pt]{fig/corr_Arora.png}
%        \caption{Arora (r=.40)}
%\end{subfigure}
\begin{subfigure}{0.49\textwidth}
    \centering
        \includegraphics[width=100pt]{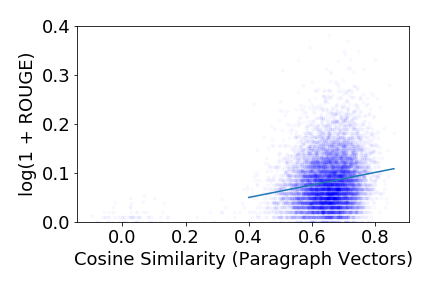}
        \includegraphics[width=100pt]{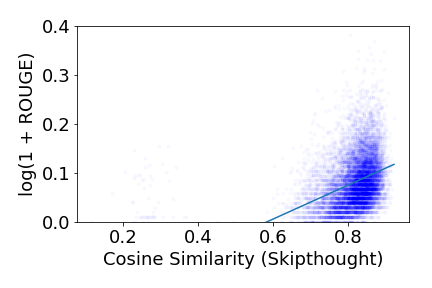}
        \caption{Paragraph Vectors (r=.17) and Skipthought (r=.33)}
        \end{subfigure}
%\begin{subfigure}{0.49\textwidth}
%    \centering
%        \includegraphics[width=150pt]{fig/corr_Skipthought.png}
%       \caption{Skipthought (r=.33)}
%\end{subfigure}
\caption{Correlation of ROUGE scores and Cosine Similarity scores per sentence. ROUGE scores transformed with $r' = log(1 + r)$, to account for zero values. Some high-leverage points have been excluded for Paragraph Vectors and Skipthought.}
\label{fig:corr}
\end{figure}
By identifying the sentences present in an optimal summarization, we show that optimal sentences have higher cosine scores, and that this effect is increased after adjusting cosine scores for word length (Fig. \ref{fig:dualcos}). However, there is a lot of overlap, implying that, although this method has some power to discern good summaries from bad summaries, the power of this method alone is not high enough to produce good summaries.

\paragraph{Regression on Vector Dimensions}
\label{sec:regression}
We calculated the isolated ROUGE score and the sentence embeddings on all four vector functions of each individual sentence in the training set. To partially eliminate the effects of the sentence's context, we subtract the corresponding document vector from all sentence vectors before regression.
Due to the large number of predictor variables, we use a Bonferroni correction, considering values significant only if they have p-values of $\frac{\alpha}{n}$, which, for $\alpha=.05$, corresponds approximately to $p < .00001$ for the skip-thought vectors, and $p < .00016$ for all other vectors.

Three dimensions are significant at this level for \textit{SIF Average} vectors. No dimensions are significant at this level for \textit{Arora} vectors, though the three significant dimensions for \textit{SIF Average} achieve noteworthy p values of .0021, .0485 and .0006. 29 dimensions are significant for \textit{Paragraph Vectors}. 5 dimensions are significant, at the much higher threshold, for \textit{Skip-Thought Vectors}. It appears that these specific dimensions correspond to aspects of a sentence that make it somehow more suited for a summary. Despite the theoretical implications of this result, the regression models do not have enough predictive power to create good summaries by themselves.

\paragraph{Example--Performance of  the Greedy Algorithm}
We explore the performance of a simple algorithm, the Greedy Algorithm. 
%The Greedy algorithm is the most successful algorithm among those we have explored. 
As its name implies, the Greedy algorithm appears to be simply an attempt at maximizing the following objective function:
\begin{equation}
f_{cos}(summary) = \overrightarrow{summary} \cdot \overrightarrow{document}.
\end{equation}
\noindent

Of course, this objective function can only be an approximation to the informally-defined criteria for good summaries. Even so, Table \ref{tab:results} suggests that the performance of the greedy algorithm is not based on the accuracy of the corresponding objective function. In particular, consider the two other strategies which try to maximize the same objective function: Brute force, and Maximum Similarity (which simply selects Greedy or Brute Force based on which one creates a summary with a higher cosine similarity). Brute Force consistently and significantly creates summaries with higher cosine similarity to the document, outperforming the Greedy selector on its objective function. By construction, the Max Similarity algorithm outperforms in cosine similarity to an even greater degree. But both of these algorithms perform much worse than the Greedy algorithm.

\begin{figure}
\centering
\includegraphics[width=.35\textwidth]{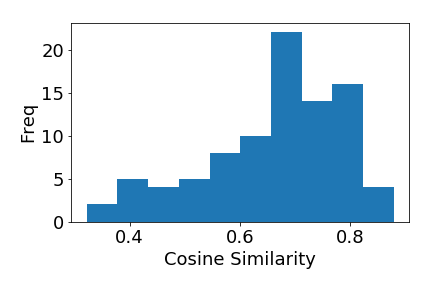}
\caption{A histogram of the cosine similarities of sentences selected by the Greedy algorithm.}
\label{fig:greedycoshist}
\end{figure}

Deeper analysis into the decisions of the Greedy algorithm reveals some reasons for this discrepancy. It appears that the good performance of the Greedy algorithm results not from the associated objective function, but by the way in which it maximizes this objective function. In particular, the Greedy algorithm selects sentences with low cosine similarity scores in a vacuum, but which increase the cosine similarity of the overall sentence (Fig. \ref{fig:greedycoshist}). 

To understand why this is true, we consider the step-by-step behavior of the Greedy algorithm. The first choice of the greedy algorithm is simple: it chooses the sentence with maximum cosine similarity to the document vector: 
%\begin{equation}
$\bar{s}_1 = \argmax_{\bar{s} \in S} \bar{s} \cdot \bar{d}.$
%\end{equation}
Recall that all vectors have unit-length, so cosine similarity is equivalent to the dot product. 

To select the second vector, the greedy algorithm is maximizing the following equation: 
\begin{align}
\bar{s}_2 &= \argmax_{\bar{s} \in S'} (\frac{\bar{s} + \bar{s}_1}{\lVert \bar{s} + \bar{s}_1\rVert}) \cdot \bar{d} \nonumber \\&= \argmax_{\bar{s} \in S'} \frac{\bar{d} \cdot \bar{s}_1 + \bar{d} \cdot \bar{s}}{\sqrt{1 + \bar{s}_1 \cdot \bar{s}}}  &
\end{align}
{where $S'$ is the set of remaining sentences.\nonumber \footnotemark}

%\footnotetext{Note that the results reported above do not represent the Greedy algorithm averaging together the vectors, %though the difference is minimal for SIF Average and Arora vectors(see Section \ref{sec:docvec} for more information)}

This equation consists of three parts: $\bar{d} \cdot \bar{s}_1$ (a constant wrt. $\bar{s}$), $\bar{d} \cdot \bar{s}$, which is simply the salience of a sentence measured by cosine similarity, and the denominator, which is essentially a measure of redundancy. Not only does this simple metric lead to a `natural' penalty for redundancy, it performs better than our handcrafted redundancy penalties. The way this algorithm scales when picking the $i^{th}$ sentence is particularity noteworthy.
\begin{align}
\bar{s}_{i + 1} &= \argmax_{\bar{s} \in S'} (\frac{\frac{1}{i + 1}\bar{s} + \frac{i}{i + 1}\bar{s}_p}{\lVert \frac{1}{i + 1}\bar{s} + \frac{i}{i + 1}\bar{s}_p\rVert}) \cdot \bar{d} \nonumber\\
&= \argmax_{\bar{s} \in S'} \frac{i(\bar{d} \cdot \bar{s}_p) + \bar{d} \cdot \bar{s}}
{\sqrt{i^2 + 1  + 2i\bar{s}_p \cdot \bar{s}}}  \label{eq:si} &
\end{align}
where  $\bar{s}_p = \frac{\sum_{j=0}^i\bar{s}_j}{\lVert \sum_{j=0}^i\bar{s}_j\rVert}$. 

\begin{figure}
\centering
\includegraphics[width=.35\textwidth]{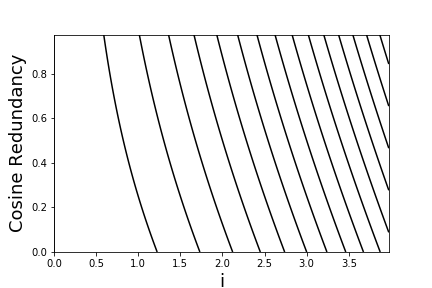}
\caption{A contour plot of the denominator of Equation \ref{eq:si}}
\label{fig:contour}
\end{figure}

As shown in Figure \ref{fig:contour}, the behavior of this function changes as $i$ increases. In particular, the function becomes more sensitive to redundancy, and less sensitive to salience, as the algorithm selects more sentences. In other words, the algorithm will first try to select important sentences, and then select sentences to fill in the gaps. This result, and the success of the resulting algorithm, has implications for balancing salience and redundancy in future summarization algorithms.

It is possible to analyze the performance of the other algorithms in a similar manner as well. 

\paragraph{Document Vector Computation}
\label{sec:docvec}
In general, there are two ways to compute a document vector. The most obvious is to pass the entire text of the document into the vector function. This has two theoretical problems. The first is that the `documents' in our algorithms are really clusters of documents, and are therefore non-coherent. The second is that Skip-thought vectors are not designed to handle text longer than a sentence. However, an alternative document vector, \textit{docvec-avg}, is formed by taking the mean of the (normalized) sentence vectors. This corresponds to treating the document as a collection of sentences, instead of a collection of words. We present a comparison of the two methods in Table \ref{tab:docvec}.

As expected, Skipthought vectors, which are not designed for text larger than a sentence, perform significantly better with the \textit{docvec-avg} strategy. More notable is the poor performance of the \textit{docvec-avg} strategy with Paragraph Vectors. The size of the performance gap here implies that Paragraph Vectors can combine information from multiple sentences in a manner more sophisticated than simple averaging.

More interesting is the performance for SIF Average and Arora vectors. For these vector functions, which are based on taking the average of words, \textit{docvec-avg} very closely resembles the simple strategy. And yet there is a small but significant performance gap. The difference between the two document vectors is the weighting. \textit{Docvec-avg}, which normalizes vectors before adding them together, removes some weighting information present in the simple strategy. In particular, the simple strategy assigns more weight to sentences with a lot of highly-weighted words. Presumably, \textit{docvec-avg}, by ignoring this weighting, leaves out useful information. This hypothesis is supported by the greater performance gap for Arora vectors, which effectively downweights certain common words and therefore could be expected to carry more information in word weightings. Similar, but much smaller, gaps exist when computing the vectors for summaries at each step in the greedy algorithm.

\paragraph{Properties of Different Sentence Embeddings}

Based on the interactions between selector functions and vector functions, as well as a variety of other pieces of data in the proceeding sections, we present a broad comparison of the properties of different sentence embedding schemes.

{\em SIF Average/Arora Vectors:}
Three selector functions perform better with both SIF Average and Arora vectors: Near Nonredundant, Greedy, and PCA. These functions seem to be united in their comparisons between the vectors of sentence embeddings (implicitly, in the case of the greedy algorithm). These selector functions correspond to the most basic test for sentence embeddings: Judging the similarity of two sentences.

The exact difference the common component removal makes is less clear. Arora vectors hold a slight performance edge for all selectors except for Near and Greedy (the Greedy algorithm loses 2 points).

{\em Paragraph Vectors:}
Two selector functions perform better with Paragraph Vectors: Near and Brute Force, which both compare sentence vectors to the document vector. The poor performance on algorithms such as Near Nonredundant suggests that Paragraph Vectors are especially poor at comparing sentence vectors to each other. This suggests that Paragraph Vectors are especially good at computing document vectors, also implied by the discussions of paragraph $e)$ above.
The other distinguishing property of Paragraph Vectors is their very high correlation in regression on the individual features.

{\em Skipthought Vectors:}
It is hard to disentangle the properties of Skipthought vectors from the high dimensionality of the pre-trained vectors we used. In general, Skipthought vectors performed poorly. They only performed better than other vector functions with one selector, Clustering, although their performance with this selector was significant. 

\section{Conclusions and Future Work}
This paper set out with the objective of finding the relationship between sentence vectors and summaries of documents. We  presented an in-depth experimental analysis of how various commonly used sentence vectors and document vectors compare in the context of summarization algorithms. 
The best performing selector, Greedy, is both very simple and based on fundamental principles of vector semantics.
Paragraph Vectors work much worse with the Clustering and Greedy algorithms, and work much better with Near.
Many combinations of selector function and vector function do not work above the level of random chance.
In general, despite their sophistication, Paragraph Vectors and Skip-Thought vectors perform worse than much more basic approaches.

In this paper, we have identified differences in different forms of sentence vectors when applied to real-world tasks. In particular, each sentence vector form seems to be more successful when used in a particular way. Roughly speaking, Arora's vectors excel at judging the similarity of two sentences while Paragraph Vectors excel at representing document vector, and at representing features as dimensions of vectors. While we do not have enough data to pinpoint the strengths of Skipthought vectors, they seem to work well in specific contexts that our work did not fully explore. These differences are extremely significant, and will likely make or break real-world applications. Therefore, special care should be taken when selecting the sentence vector method for a real-world task. 

Dealing with sentences is quite common in many natural language processing tasks. Whether it is comparing two sentences for semantic similarity or entailment of one in the other, performing extractive or abstractive summarization, translating a sentence in one language to another, or a host of other tasks, the computation of sentence vectors as an intermediate step often plays a crucial role in achieving success. Many approaches build their sentence vectors in especially designed CNN, RNN and/or attention-based models. It will be interesting and useful to extend our current work to examine the sentence vectors being produced by various architectures to analyze their suitability for various tasks. If one can establish the suitability of certain sentence encoders for certain tasks, these could be used as modules for the respective tasks in lieu of each effort building their own  sentence encoders. 

%\subsubsection*{Acknowledgments}
%This material is based upon work supported by the National Science Foundation under Grant No. 1659788 and 1359275

\balance
\bibliographystyle{IEEEtran}

\bibliography{templeton-reu17}

\end{document}